\documentclass[journal, onecolumn]{IEEEtran}

\usepackage[utf8]{inputenc}
\usepackage[T1]{fontenc}
\usepackage{amsmath,amssymb}
\usepackage{listings}
\usepackage{xcolor}
\usepackage{tabularx}
\usepackage{siunitx}
\usepackage{cite} 
\usepackage{hyperref}

\usepackage{verbatim}
\usepackage{booktabs}
\usepackage{enumitem}
\usepackage{geometry}
\usepackage{algorithm}
\usepackage{algpseudocode}
\usepackage{graphicx}
\usepackage{subcaption}
\usepackage{hyperref}
\usepackage{tikz}
\usetikzlibrary{shapes.geometric, arrows.meta, positioning, calc}

\ifCLASSINFOpdf
\else
\fi
\begin{document}

\title{HLF-FSL: A Decentralized Federated Split Learning Solution for IoT on Hyperledger Fabric}

\author{
\IEEEauthorblockN{Carlos Beis-Penedo\IEEEauthorrefmark{1}, 
Rebeca P. Díaz-Redondo\IEEEauthorrefmark{1}, \\ 
Ana Fernández-Vilas\IEEEauthorrefmark{1}, Manuel Fernández Veiga\IEEEauthorrefmark{1}
Francisco Troncoso-Pastoriza\IEEEauthorrefmark{1}\IEEEauthorrefmark{2}\\
}
\IEEEauthorblockA{\IEEEauthorrefmark{1}\textit{atlanTTic Research Center, Information \& Computing Lab (ICLAB)} \\
\textit{Telecommunication Engineering School, Universidade de Vigo, Vigo, 36310, Spain} \\
\{cbeis,rebeca,avilas, mveiga\}@det.uvigo.es, \{ftroncoso\}@cud.uvigo.es}
\IEEEauthorblockA{\IEEEauthorrefmark{2}\textit{CUD Escuela Naval Militar} \\
\textit{Universidade de Vigo / Escuela Naval Militar}\\
Vigo / Marín, Spain}
}

\maketitle

\begin{abstract}
Collaborative machine learning in sensitive domains demands scalable, privacy‐preserving solutions for enterprise deployment. Conventional Federated Learning (FL) relies on a central server—introducing single points of failure and privacy risks—while Split Learning (SL) partitions models for privacy but scales poorly due to sequential training. We present a decentralized architecture that combines Federated Split Learning (FSL) with the permissioned blockchain Hyperledger Fabric (HLF) that we have coined as HLF-FSL.  Our chaincode orchestrates FSL’s split‐model execution and peer‐to‐peer aggregation without any central coordinator, leveraging HLF’s transient fields and Private Data Collections (PDCs) to keep raw data and model activations private.  On CIFAR-10 and MNIST benchmarks, HLF-FSL matches centralized FSL accuracy while reducing per-epoch training time compared to Ethereum-based works. Performance and scalability tests show minimal blockchain overhead and preserved accuracy, demonstrating enterprise-grade viability.
\end{abstract}


\IEEEpeerreviewmaketitle

\section{Introduction}
\label{sec:introduction}

The rise of collaborative machine learning unlocked new possibilities in sensitive domains like healthcare and finance, but it still presents a fundamental challenge: how to train models on distributed data without compromising privacy or scalability \cite{poirot2019split} for the dataholders. Standard paradigms like Federated Learning (FL) \cite{mcmahan2017communication} offer a partial solution by keeping raw data local, but their dependence on a central server introduces a bottleneck and a target for privacy attacks~\cite{fredrikson2015model, zhu2019deep, zhang2024tackling}. Opposite, Split Learning (SL) \cite{Gupta2018} enhances privacy by splitting the model, but its sequential behavior creates a performance bottleneck that slows down large-scale deployment\cite{zhang2023privacy}, also suffering from activation-based data leakage threats~\cite{pham2023data}. Bridging this gap requires an architecture that can get both parallel processing and partitioned privacy.

To address this limitations, Federated Split Learning (FSL)~\cite{Thapa2022} merges these paradigms, enabling parallel client execution while maintaining model partitioning. However, existing FSL implementations depend on centralized servers to host critical model segments and coordinate aggregation, perpetuating trust assumptions and availability bottlenecks. The challenges, however, run deeper than just replacing a single server. A truly robust and enterprise-grade solution must address a confluence of unresolved problems. Firstly, even in a split model, the intermediate activations and gradients exchanged between participants are not inherently safe; they remain vulnerable to sophisticated inference and reconstruction attacks that can compromise client data privacy. Secondly, as models grow in complexity and size, the practical issue of verifiably managing and aggregating large parameter sets in a distributed setting without a trusted central authority becomes a significant logistical and security hurdle. A comprehensive framework must therefore not only decentralize coordination but also provide granular, verifiable mechanisms for protecting transient computational data and managing large-scale off-chain assets.

This work resolves this situation by integrating FSL with Hyperledger Fabric (HLF)~\cite{androulaki2018hyperledger}, a permissioned enterprise blockchain framework known for its modularity, fine-grained access control, and private transaction capabilities. We reimagine the blockchain not as a computational engine, but as a decentralized and auditable coordination substrate. Our system uses HLF's smart contracts (chaincode) to orchestrate the entire FSL workflow, from model management to the aggregation of updates, without a central coordinator. To shield sensitive intermediate data from on-ledger persistence and inference attacks, we leverage HLF's transient fields for ephemeral data exchange and Private Data Collections (PDCs) to manage access-controlled references. For managing large model parameters, our architecture stores them off-chain while using the blockchain to immutably log and verify their cryptographic hashes, ensuring integrity without bloating the ledger. This design employs the blockchain to enforce governance, manage participant identities, and create a verifiable log of all training activities, distributing trust across the consortium while keeping the intensive machine learning computations off-chain.

This paper makes three key contributions. We (1) design the first FSL architecture that eliminates the central coordinator by using HLF for decentralized orchestration;(2) introduce an enhancing privacy implementation through the use of transient fields and Private Data Collections to shield data from on-ledger persistence and unauthorized access; and (3) develop a novel chaincode-driven protocol for managing the FSL lifecycle, including a verifiable, client-executed aggregation process. Experiments on CIFAR-10 and MNIST validate HLF-FSL's viability: the system matches centralized FSL accuracy  while reducing per-epoch training time compared to prior Ethereum-SL~\cite{beis24} or FSL~\cite{Batool2024} implementations. Scalability tests demonstrate efficient operation with an elevated number of simultaneous clients, as Fabric's modular design minimizes blockchain overhead. An open-source prototype confirms the practicality of this architecture for enterprise deployments, where regulatory compliance and operational governance are paramount.

In the following sections of the paper we can find background about FSL and Hyperledger Fabric (Section~\ref{sec:background}), comparison and reviews with the state-of-the-art related works on the field (Section~\ref{sec:related}), the complete proposed HLF-FSL architecture and methodology (Section~\ref{sec:methodology}), all the experimental setup (Section~\ref{sec:experimental_setup}), the results obtained through all the testing (Section~\ref{sec:results}) and, finally, the conclusion of the paper and the disccussion of future work (Section~\ref{sec:conclusion}).

\section{Background}
\label{sec:background}

Collaborative machine learning aims to use data from multiple sources without centralizing raw data, addressing privacy concerns inherent to traditional approaches. This section reviews Federated Split Learning, the foundational distributed learning paradigm relevant to our work. We then discuss the role of blockchain technology, particularly Hyperledger Fabric, in providing decentralized trust and coordination for this paradigm.

\subsection{Federated Split Learning}

\begin{figure*}[tbp!]
  \centering
  \includegraphics[width=0.95\textwidth]{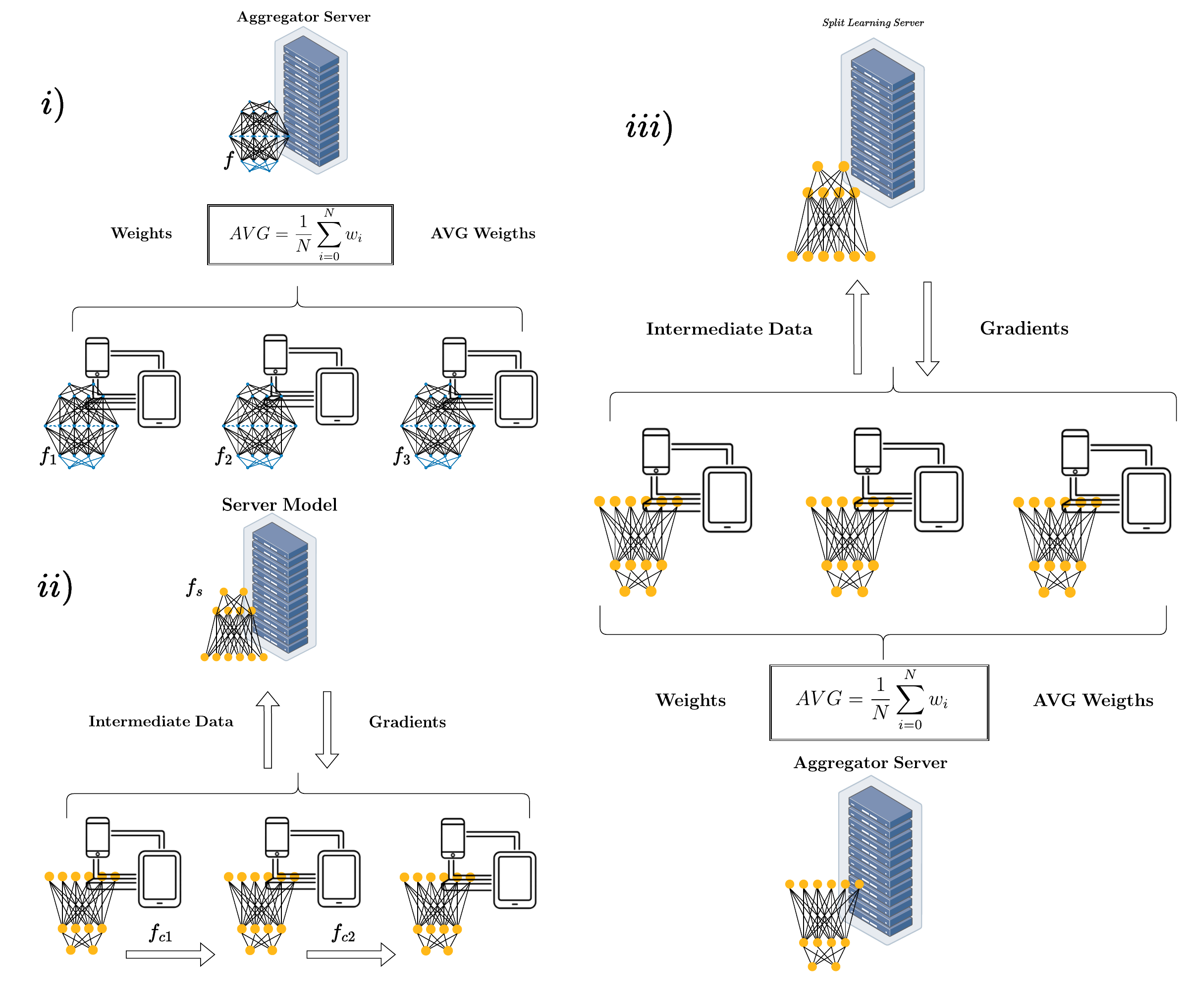}
  \caption{Overview of distributed learning architectures: (i) Federated Learning, where clients share model updates with a central aggregator; (ii) Split Learning, where the model is split and clients share intermediate activations sequentially with a server; and (iii) Federated Split Learning, combining model splitting with parallel client processing and federated aggregation of client-side updates.}
  \label{fig:decentralized_architecture}
\end{figure*}

FSL, shown in Fig.~\ref{fig:decentralized_architecture}.(iii), also known as SplitFed~\cite{Thapa2022}, merges the strengths of FL (Fig.~\ref{fig:decentralized_architecture}.(i)) and SL (Fig.~\ref{fig:decentralized_architecture}.(ii)), creating a hybrid architecture. It retains SL's model splitting architecture ($f_c$ on clients, $f_s$ on a server) but integrates FL's parallel client processing. In a typical FSL round $t$, a set of participating clients $S_t$ is selected. For each client $i \in S_t$ with its local dataset $D_i$ containing data samples $\mathbf{X}_i$, the process is as follows:
\begin{enumerate}
    \item All participating clients $i$ \textit{in parallel} compute activations $\mathbf{Z}_i^{(t)} = f_c(\mathbf{X}_i; \mathbf{w}_c^{(t)})$ using the current global client-side model $\mathbf{w}_c^{(t)}$ and send them to the server.
    \item The server processes these activations through its segment $f_s(\mathbf{Z}_i; \mathbf{w}_s^{(t)})$, computes the loss $L_i$ for each client's batch, and calculates gradients with respect to its own weights $\nabla_{\mathbf{w}_s} L_i$ and the received activations $\nabla_{\mathbf{Z}_i} L_i$.
    \item The server updates its own parameters $\mathbf{w}_s$ based on aggregated gradients, typically using a learning rate $\eta_s$ ($\mathbf{w}_s^{(t+1)} = \mathbf{w}_s^{(t)} - \eta_s \sum_i \nabla_{\mathbf{w}_s} L_i$).
    \item The server sends gradients $\nabla_{\mathbf{Z}_i} L_i$ back to the respective clients.
    \item Each client $i$ performs backpropagation through $f_c$ to compute gradients for its local model, $\nabla_{\mathbf{w}_c} L_i$, and obtains temporary updated client-side parameters $\tilde{\mathbf{w}}_{c,i}^{(t+1)}$.
    \item Periodically, these temporary client-side parameters $\tilde{\mathbf{w}}_{c,i}^{(t+1)}$ are aggregated (typically using FedAvg, weighted by the size of each client's dataset $|D_i|$) to produce the next global client-side model $\mathbf{w}_c^{(t+1)}$:
    \begin{equation}
    \mathbf{w}_c^{(t+1)} = \sum_{i \in S_t} \frac{|D_i|}{\sum_{j \in S_t} |D_j|} \tilde{\mathbf{w}}_{c,i}^{(t+1)}
    \label{eq:fsl_client_agg} 
    \end{equation}
    This new $\mathbf{w}_c^{(t+1)}$ is distributed to clients for the subsequent round.
\end{enumerate}
FSL enhances privacy compared to FL (server doesn't see client gradients directly) and improves efficiency over SL (clients compute in parallel). However, standard FSL still often relies on a central server entity to manage $\mathbf{w}_s$, process activations, and coordinate the federated aggregation of $\mathbf{w}_c$. This centralization remains a potential bottleneck and point of trust, motivating the exploration of blockchain technology for further decentralization and enhanced security.

\subsection{Blockchain for Decentralized Trust and Coordination}
\label{subsec:blockchain} 

Distributed Ledger Technologies (DLTs), particularly Blockchain, provide a fundamentally different model for trust and coordination compared to centralized systems~\cite{zheng2018blockchain}. Blockchain acts as a shared, immutable ledger maintained by a network of participants through consensus protocols. This eliminates the need for a single trusted intermediary, enabling verifiable recording of transactions (such as model updates or aggregation results) and automated execution of agreements via smart contracts~\cite{khan2021blockchain}.

While early explorations used public, permissionless blockchains like Ethereum for FL~\cite{beis24}, they often encountered practical limitations unsuitable for iterative ML workloads~\cite{nguyen2024wait}. The low transaction throughput and high confirmation latency inherent to consensus mechanisms like Proof-of-Work are one major obstacle. A second, more critical issue for machine learning is the high cost due to every transaction that requires mandatory mmonetary payment, knownn as gas fee, tom compensate network validator for computational resources. This is designed to prevent network abuse and, since in decentralized learning processes every model update constitutes a transaction, the cumulative gas become prohibitevely expensive, rendering them impractical for real-world scenarios. Furthermore, other fundamental concern identified is that data on these blockchains are publicly available by default. This could discourage user participation in decentralized federated learning protocols, as data protection is typically the primary motivating factor for FL~\cite{kairouz2019advances}.

These limitations of public blockchains have motivated a necessary shift towards permissioned blockchain frameworks, which are specifically designed to offer the governance, privacy, and performance characteristics required for real world use cases collaborative learning.

\subsection{Hyperledger Fabric: A Permissioned Blockchain for Enterprise Solutions}
\label{subsec:hlf} 

Hyperledger Fabric~\cite{androulaki2018hyperledger} is an open-source, permissioned blockchain framework designed for enterprise use cases. Its modular architecture and specific features address many limitations of public blockchains, making it a strong candidate for develop robust collaborative ML systems like FSL. Key relevant features include:
\begin{itemize}
    \item \textbf{Permissioned Network \& MSP:} Fabric operates among known, authenticated participants managed by a Membership Service Provider (MSP). This allows for fine-grained access control, crucial for managing roles (clients, servers, auditors) and ensuring only authorized entities interact with the FSL process and data.
    \item \textbf{Execute-Order-Validate Transaction Flow:} Fabric separates transaction execution (endorsement by peers running chaincode) from transaction ordering (by a pluggable consensus service, e.g., Raft) and final validation (by committing peers) \cite{ruan2020transactional}. This enables parallel execution, higher throughput, and faster finality compared to order-execute models like Ethereum's.
    \item \textbf{Chaincode (Smart Contracts):} Application logic is implemented in chaincode (written in Go, Java, etc.). Chaincode can automate FSL coordination steps, perform on-chain aggregation (within limits), manage model registries, and enforce validation rules in a verifiable manner.
    \item \textbf{Channels and Privacy Features:} Fabric Channels allow private subnetworks within the larger network, isolating data and transactions for specific collaborations. Furthermore, Private Data Collections (PDCs) \cite{hyperledgerPDC} and Transient Fields allow sensitive data (like FSL activations or gradients) to be shared only among authorized peers or used during chaincode execution without being permanently recorded on the main ledger, significantly enhancing privacy.
    \item \textbf{Performance and Scalability:} Using efficient consensus protocols like Raft (Crash Fault Tolerant), Fabric achieves much higher throughput (potentially thousands of transactions per second) and lower latency (sub-second finality) than PoW-based chains, making it viable for the iterative communication rounds required by FSL.
\end{itemize}
These features—particularly permissioned governance, enhanced privacy mechanisms, and superior performance—position HLF as a suitable technological foundation for building the secure, scalable, and enterprise-ready FSL system we propose in this work. 

\section{Related Work}
\label{sec:related}

The proposed integration of Federated Split Learning (FSL) with Hyperledger Fabric (HLF) builds upon advancements in distributed machine learning, blockchain technology, and enterprise system design. This section critically examines pertinent literature to contextualize our framework's novelty and contributions, highlighting how HLF addresses key limitations in existing approaches. To synthesize these diverse approaches and clearly position our contribution, Tables~\ref{tab:related_works_part1} and \ref{tab:related_works_part2} provide a detailed comparative analysis where our proposed HLF-FSL system is included alongside existing frameworks to clearly delineate its architectural and functional advantages.

\subsection{The Evolution from FL and SL to Federated Split Learning}
Split Learning (SL) fundamentally altered distributed training by partitioning neural networks between clients and a server, enabling collaborative model training without direct raw data sharing \cite{Gupta2018}. In SL, clients process initial model layers and transmit only intermediate activations, offering inherent privacy benefits \cite{poirot2019split, Gupta2018}. Foundational works validated SL's effectiveness, particularly in sensitive domains like healthcare where raw data cannot be pooled \cite{poirot2019split}. However, SL also exposed limitations: the potential for information leakage from activations necessitated countermeasures such as noise addition or activation truncation, revealing an inherent privacy-utility trade-off. More critically for scalability, SL's canonical protocol, where the server typically processes clients sequentially, creates a significant performance bottleneck as participant numbers grow \cite{Thapa2022}.

FSL emerged to synthesize SL's privacy model with Federated Learning's (FL) parallel client processing capabilities. Thapa et al.'s SplitFed architecture \cite{Thapa2022} demonstrated that this hybrid could achieve faster convergence than vanilla FL by leveraging parallel client-side ($f_c$) computations while still splitting the model. However, SplitFed and subsequent FSL variants, such as group-based SFL for wireless networks \cite{zhang2023split} and edge-assisted U-shaped FSL for IoT \cite{zhang2025edge}, often retained a centralized server or trusted intermediate aggregators (e.g., edge servers or a central cloud in hierarchical FSL for smart grids \cite{li2024introducing}). While these architectures improved latency and resource efficiency in constrained settings, they did not fully resolve issues associated with single points of failure or the need for complete trust in these intermediate coordinating entities. Furthermore, the core challenge of potential information leakage from intermediate activations, even in split models \cite{pasquini2022eluding, abuadbba2020can}, persisted, underscoring the need for more robust, decentralized, and verifiably secure coordination frameworks.

\subsection{Blockchain for Decentralized Learning: From Public to Permissioned}
Blockchain technology offers a solution for decentralized trust and verifiable record-keeping, making it a natural candidate for enhancing the security and transparency of distributed ML systems. Early frameworks, such as BAFFLE by Ramanan and Nakayama (2020) \cite{ramanan2020baffle}, demonstrated the feasibility of replacing the central FL server with Ethereum smart contracts, achieving an aggregator-free setup. However, these public blockchain implementations, including others like OpenFL by Wahrstätter et al. (2024) \cite{wahrstatter2024openfl}, consistently faced significant performance penalties. The low transaction throughput, high confirmation latency (due to Proof-of-Work consensus), and expensive transaction costs (gas fees) inherent to platforms like Ethereum make them generally unsuitable for the frequent, low-latency communication rounds required by iterative ML \cite{nguyen2024wait}.

These limitations encouraged the exploration of permissioned blockchains for FL. Consortium chains, as discussed by Wang et al. (2023) \cite{wang2023enhancing}, can incorporate on-chain reputation mechanisms to enhance trustworthiness among known participants, but introducing notable communication overhead. Works like Putra et al. have explored permissioned Proof-of-Authority blockchains with differential privacy for industrial IoT, emphasizing efficiency and security in these settings~\cite{Proof-of-authority-Adi}. Conceptual work on Corda-based FL also highlights the suitability of permissioned designs for private data exchange and secure computation in enterprise settings \cite{shitharth2023federated}. As synthesized in surveys of Blockchain FL schemes (e.g., Lu et al., 2022 \cite{singh2024systematic}), we can detect a common trade-off: fully on-chain processing maximizes decentralization and auditability but often bottlenecks training efficiency. Hybrid designs, which perform most computations off-chain while using the blockchain for coordination and verification, typically improve speed but require careful consideration of trust assumptions for off-chain components. Similarly, new frameworks combine blockchain with advanced privacy-enhancing technologies like homomorphic encryption to protect sensitive healthcare data, underscoring the trend toward robust, privacy-first architectures~\cite{FIRDAUS2025101579}. Prior blockchain-ML systems, while improving trust through immutable logging (e.g. Oikonomou et al. \cite{oikonomou2021hyperledger} using Fabric for FL), frequently overlooked critical requirements. These include granular identity management, fine-grained privacy controls that go beyond simply withholding raw data (like protecting intermediate computational artifacts), and seamless integration with existing data governance frameworks. These specific needs are central to our adoption of HLF.

\subsection{The Research Gap - Decentralized FSL}
\label{subsec:hlf_for_fsl_revised}

While HLF's general suitability for enterprise applications is well-established \cite{androulaki2018hyperledger}, its specific application to FSL to overcome existing limitations remains largely unexplored. Prior HLF-based Federated Learning (FL) solutions, such as FedChain \cite{wu2024blockchain}, have utilized HLF for decentralized aggregation and auditability. However, these works did not address the more complex data flows of FSL nor leverage HLF's advanced privacy features for the nuanced challenge of protecting intermediate computational artifacts.

The migration from public to permissioned blockchains for ML is motivated by significant performance gains in throughput and latency, a fact supported by benchmarks \cite{dinh2017blockbench, xu2021latency} and observed in practice when moving from Ethereum \cite{beis24, Batool2024} to HLF. Despite these advantages, a clear research gap remains for an architecture that can:
\begin{enumerate}
    \item Orchestrate the FSL workflow \cite{Thapa2022} without a centralized server dependency for coordination and aggregation.
    \item Protect sensitive intermediate data—which remains vulnerable to inference attacks even in a split setting \cite{pasquini2022eluding, abuadbba2020can}—using granular, off-ledger privacy mechanisms native to the blockchain platform.
    \item Provide a flexible and verifiable management strategy for model parameters, whether they are small enough for on-chain private storage or require referencing large-scale off-chain systems like IPFS.
\end{enumerate}

Our work is the first to address this specific confluence of challenges. By strategically employing HLF's transient fields for ephemeral data, Private Data Collections for access-controlled hash management, and an event-driven chaincode to orchestrate a fully client-led aggregation, we demonstrate a system that achieves a higher degree of decentralization, verifiable privacy, and enterprise-readiness than existing approaches. To synthesize these diverse approaches and clearly position our contribution, Tables~\ref{tab:related_works_part1} and \ref{tab:related_works_part2} provides a comparative analysis across key architectural, decentralization, privacy, and enterprise features.

\begin{table*}[!tbp]
\centering
\caption{Comparative analysis of related works (Part 1: Architecture and Decentralization)}
\label{tab:related_works_part1}
\resizebox{\textwidth}{!}{
\begin{tabular}{@{}lllll@{}}
\toprule
\textbf{Reference} & \textbf{ML Arch. Focus} & \textbf{Blockchain Used} & \textbf{Decentralization Level} \\
\midrule
Poirot \textit{et al.} (2019) \cite{poirot2019split} & Split Learning                       & None                          & Low (Central Server) \\
Gupta \& Raskar (2018) \cite{Gupta2018}         & Split Learning                       & None                          & Low (Central Server) \\
Thapa \textit{et al.} (2022) \cite{Thapa2022}  & FedSplit Learning             & None                          & Medium (Central Coord. Server) \\
Beis-Penedo \textit{et al.} (2024) \cite{beis24} & Split Learning                       & Ethereum                      & High (Public Blockchain) \\
Zhang \textit{et al.} (2023) \cite{zhang2023split} & Split Learning + Grouping            & None                          & Medium (Grouped Servers) \\
Zhang \textit{et al.} (2025) \cite{zhang2025edge} & FedSplit Learning & None & Medium (Edge Coordination) \\
Li \textit{et al.} (2024) \cite{li2024introducing} & Split Learning                       & None                          & Medium (Centralized Cloud Agg.) \\
Ramanan \& Nakayama (2020) \cite{ramanan2020baffle} & Federated Learning                         & Ethereum                      & High (Aggregator-Free) \\
Wahrstätter \textit{et al.} (2024) \cite{wahrstatter2024openfl} & Federated Learning                         & Ethereum                      & High (Aggregator-Free) \\
Wang \textit{et al.} (2023) \cite{wang2023enhancing} & Federated Learning                         & Permissioned BC               & High (Consortium) \\
Shitharth \textit{et al.} (2023) \cite{shitharth2023federated} & Federated Learning                         & Corda                         & High (Consortium) \\
Wu \textit{et al.} (2024) \cite{wu2024blockchain} & Federated Learning                         & Hyperledger Fabric            & High (Consortium) \\
Batool \textit{et al.} (2022) \cite{Batool2024}   & FedSplit Learning             & Ethereum                      & High (Aggregator-Free) \\
\midrule
\textbf{This work} & \textbf{FedSplit Learning} & \textbf{Hyperledger Fabric} & \textbf{High (Chaincode Coord.)} \\
\bottomrule
\end{tabular}
}
\end{table*}

\begin{table*}[!tbp]
\centering
\caption{Comparative analysis of related works (Part 2: Aggregation, Privacy, and Enterprise Features)}
\label{tab:related_works_part2}

\begin{tabular*}{\textwidth}{@{\extracolsep{\fill}}llll@{}}
\toprule
\textbf{Reference} & \textbf{Aggregation Method} & \textbf{Privacy Techniques} & \textbf{Enterprise Features} \\
\midrule
Poirot \textit{et al.} (2019) \cite{poirot2019split} & Centralized Server                       & Model Splitting               & None \\
Gupta \& Raskar (2018) \cite{Gupta2018}         & Centralized Server                       & Model Splitting               & None \\
Thapa \textit{et al.} (2022) \cite{Thapa2022}  & Centralized FedAvg Server                & Model Splitting               & None \\
Beis-Penedo \textit{et al.} (2024) \cite{beis24} & Centralized Server                       & Model Splitting               & Smart Contracts \\
Zhang \textit{et al.} (2023) \cite{zhang2023split} & Grouped Servers                          & Local Encoding               & None \\
Zhang \textit{et al.} (2023) \cite{zhang2025edge} & Edge Aggregation                        & Local Encoding               & None \\ 
Li \textit{et al.} (2024) \cite{li2024introducing} & Centralized Cloud                        & Local Encoding               & None \\
Ramanan \& Nakayama (2020) \cite{ramanan2020baffle} & On-chain Smart Contract (FedAvg)         & None Mentioned                & None \\
Wahrstätter \textit{et al.} (2024) \cite{wahrstatter2024openfl} & On-chain Smart Contract (FedAvg)         & Smart Contract Logic         & None \\
Wang \textit{et al.} (2023) \cite{wang2023enhancing} & On-chain Aggregation                    & Reputation System             & Basic Consortium \\
Shitharth \textit{et al.} (2023) \cite{shitharth2023federated} & Off-chain Concept                       & Permissioned Access           & Limited Enterprise \\
Wu \textit{et al.} (2024) \cite{wu2024blockchain} & Chaincode (Coordination)                & Ledger Auditing               & Basic Chaincode \\
Batool \textit{et al.} (2022) \cite{Batool2024}   & On-chain Smart Contract                   & On-chain Privacy             & None \\
\midrule
\textbf{This work} & \textbf{Chaincode-Coordinated} & \textbf{Transient Fields, PDCs,} & \textbf{MSP, Channels,} \\
& \textbf{Decentralized FedAvg} & \textbf{Model Splitting} & \textbf{Endorsement Pol., Audit} \\
\bottomrule
\end{tabular*}
\end{table*}

\section{Methodology}
\label{sec:methodology}

This section details the proposed HLF-FSL platform, outlining the key design challenges addressed, the system architecture, the specific chaincode protocols designed for orchestration, and the inherent privacy mechanisms enabled by the synergy between FSL and HLF.

\subsection{Key Design Challenges for HLF-FSL}
\label{subsec:design_challenges}
Building a robust FSL platform requires addressing several key design challenges that go beyond standard FSL implementations:
\begin{itemize}
    \item \textbf{Decentralizing Aggregation and Coordination.} A primary goal is to move beyond architectures reliant on a single, central FSL server for both server-side model ($f_s$) execution and client-side model ($f_c$) aggregation. Our approach aims to distribute these coordination and computation tasks using HLF's decentralized trust mechanisms.
    \item \textbf{Ensuring Privacy of Intermediate Data in a Blockchain Setting.} Intermediate activations ($z_i$) and their corresponding gradients ($\nabla_{\mathbf{z}_i} L_i$) are core to SL and FSL but are sensitive. The challenge lies in facilitating their exchange between clients and the server entity without persisting them on the main ledger, while still maintaining an auditable record of interaction.
    \item \textbf{Verifiably Managing Off-Chain State.} Large ML model parameters ($\tilde{\mathbf{w}}_{c,i}, \bar{\mathbf{w}}_c$) are often too large for direct on-chain storage. The challenge is to store these off-chain (e.g., client-managed storage, IPFS) while using the blockchain to verifiably manage references (hashes) and control access to these parameters.
    \item \textbf{Orchestrating Client Aggregation.} If clients are to perform the federated aggregation themselves, a clear, verifiable, and secure protocol is needed to coordinate this multi-party computation, including how clients discover necessary peer model data and how they reach consensus on the new global model.
\end{itemize}

The proposed architecture and chaincode design, detailed below, are engineered to address these specific challenges.

\subsection{HLF-FSL System Architecture}

\begin{figure}[tbp!] 
  \centering
  \includegraphics[width=\linewidth]{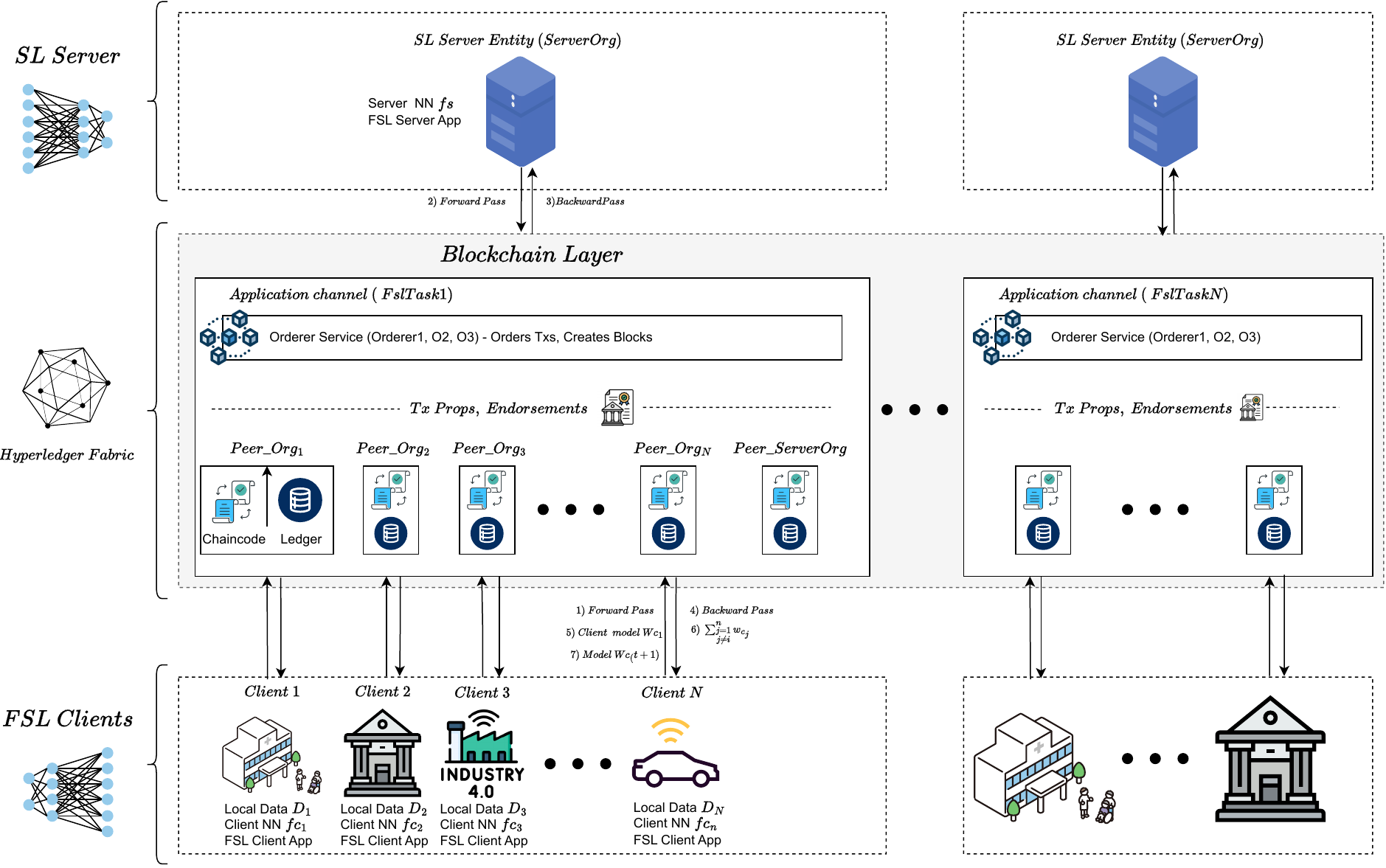} 
  \caption{Global overview of the Hyperledger Fabric Federated Split Learning (HLF-FSL) architecture, illustrating the roles of Clients, Server Entity, and the HLF Network with its components (Peers, Orderers, Ledger, PDCs, Chaincode). Arrows indicate primary data and control flows.}
  \label{fig:architecture}
\end{figure}

The HLF-FSL architecture integrates Federated Split Learning with Hyperledger Fabric to create a decentralized, verifiable, and privacy-preserving collaborative training environment. The architecture has  three core types of participants, each typically representing a distinct Fabric Organization with its own MSP: Clients (data owners, who run $f_c$ and collaboratively perform aggregation), a Server Entity (managing $f_s$), and the HLF network (peers, orderers, chaincode, ledger with PDCs), which facilitates trusted coordination and state management. An Admin Organization is also part of the consortium for policy definition and auditability. As shown conceptually in Figure~\ref{fig:architecture}, the core principle is off-chain ML computation orchestrated and verified on-chain.

\textbf{Clients} are typically enterprise entities or devices possessing private datasets. Each client trains its local sub-model ($f_C$) to generate intermediate activations ($z_i$) and updates its parameters $\tilde{\mathbf{w}}_{c,i}$ based on gradients from the Server Entity. Critically, clients also collectively perform the federated aggregation computation off-chain, using parameter hashes retrieved from the HLF network to fetch actual parameters, compute the global average $\bar{\mathbf{w}}_c$, and submit the resulting hash for on-chain recording. Fabric's MSP authenticates all client actions. 

\textbf{Server Entities} host and execute the server-side model component $f_s$. They process activations $z_i$ received from clients (data retrieved using hashes stored in PDCs, with raw data primarily passed via transient fields or IPFS) and return corresponding gradients $\nabla_{\mathbf{z}_i} L_i$ (again, using transient fields/IPFS, with hashes stored in PDCs). The Server Entity updates its own parameters $\mathbf{w}_s$ entirely off-chain. It does not participate in the aggregation of client-side models ($\mathbf{w}_c$). 

\textbf{HLF Network} consists of peer nodes (maintaining the ledger, PDCs with CouchDB, and chaincode) and orderers (Raft consensus). Each Client, Server, and Admin Organization runs peers. The network manages identities (MSP), orders transactions, maintains the immutable ledger (storing metadata, PDC references, and hashes of off-chain data), manages PDCs (storing only hashes with policies such as `OR('ClientMSP.member', 'ServerMSP.member', 'AdminMSP.member')` for intermediate data hashes, and more specific policies for model parameter hashes), executes chaincode for orchestration and facilitates communication via events. \\

Decentralized coordination is primarily chaincode-driven, a design that replaces the centralized server dependency inherent in frameworks like SplitFed \cite{Thapa2022}. Our architecture employs a hybrid on-chain/off-chain data management strategy to ensure both privacy and scalability. Sensitive, transient data like intermediate activations ($z_i$) and gradients ($\nabla_{\mathbf{z}_i} L_i$) are handled primarily through HLF's transient fields. For larger artifacts exceeding these limits, data is stored in a distributed off-chain system like IPFS, with only its immutable hash (CID) being passed through the transaction.

A similar dual strategy applies to model parameters ($\tilde{\mathbf{w}}_{c,i}, \bar{\mathbf{w}}_c$). For models adhering to HLF's state database size recommendations, parameters can be stored directly within Private Data Collections (PDCs). For massive models, parameters are stored off-chain (e.g., in client-controlled storage or IPFS), with their immutable hashes recorded and managed via PDCs. This flexible approach offers a more granular and practical privacy model for diverse model sizes. In all cases, the on-chain ledger acts as the immutable source of truth for these references (hashes), guaranteeing data integrity and providing a verifiable audit trail for all operations.\\

\begin{figure*}[tbp!]
  \centering
  \includegraphics[width=0.95\textwidth]{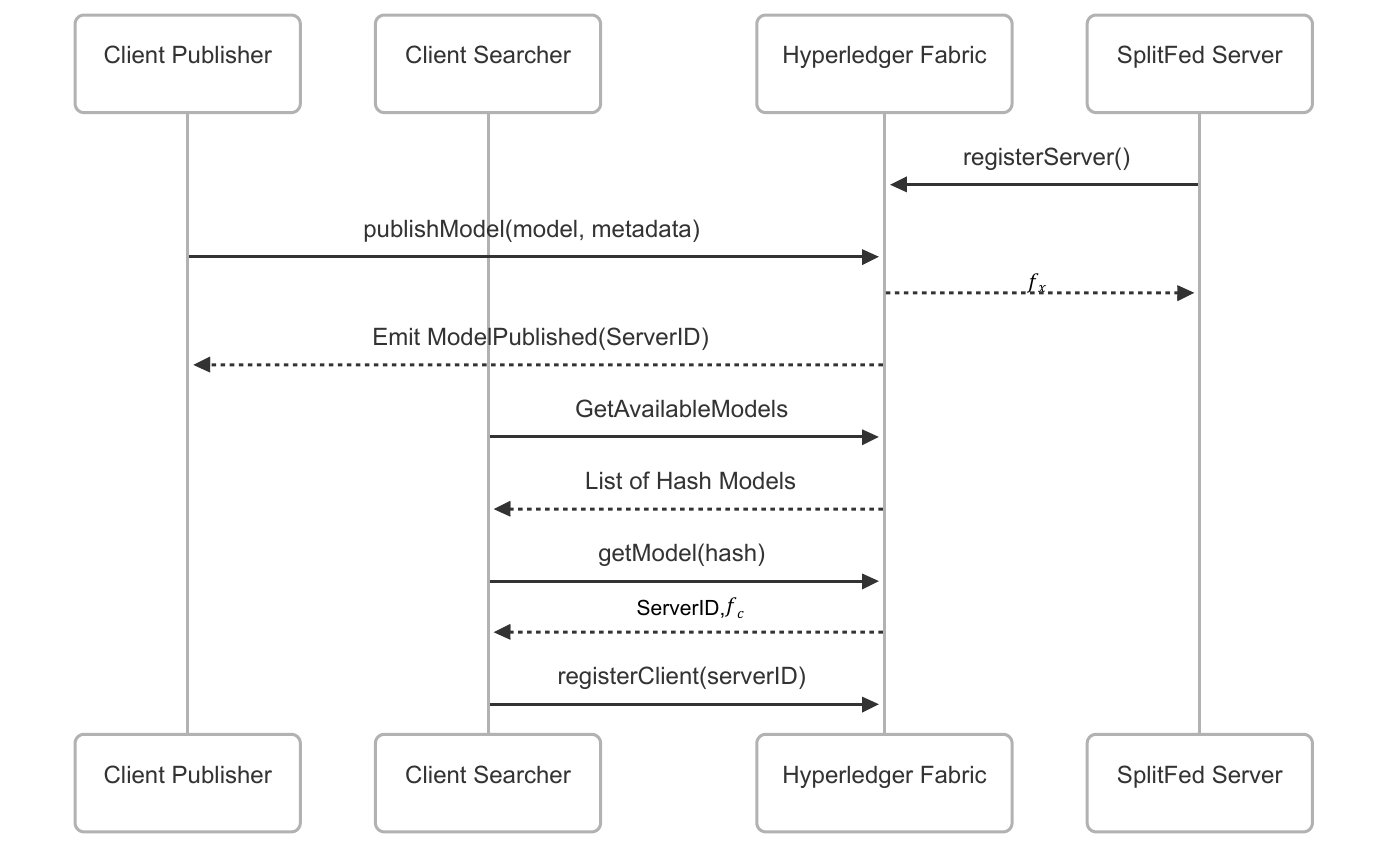}
  \caption{Model lifecycle workflow encompassing server registration, model publication, and client binding}
  \label{fig:model-lifecycle}
\end{figure*}

\begin{figure*}[tbp!]
  \centering
  \includegraphics[width=0.95\textwidth]{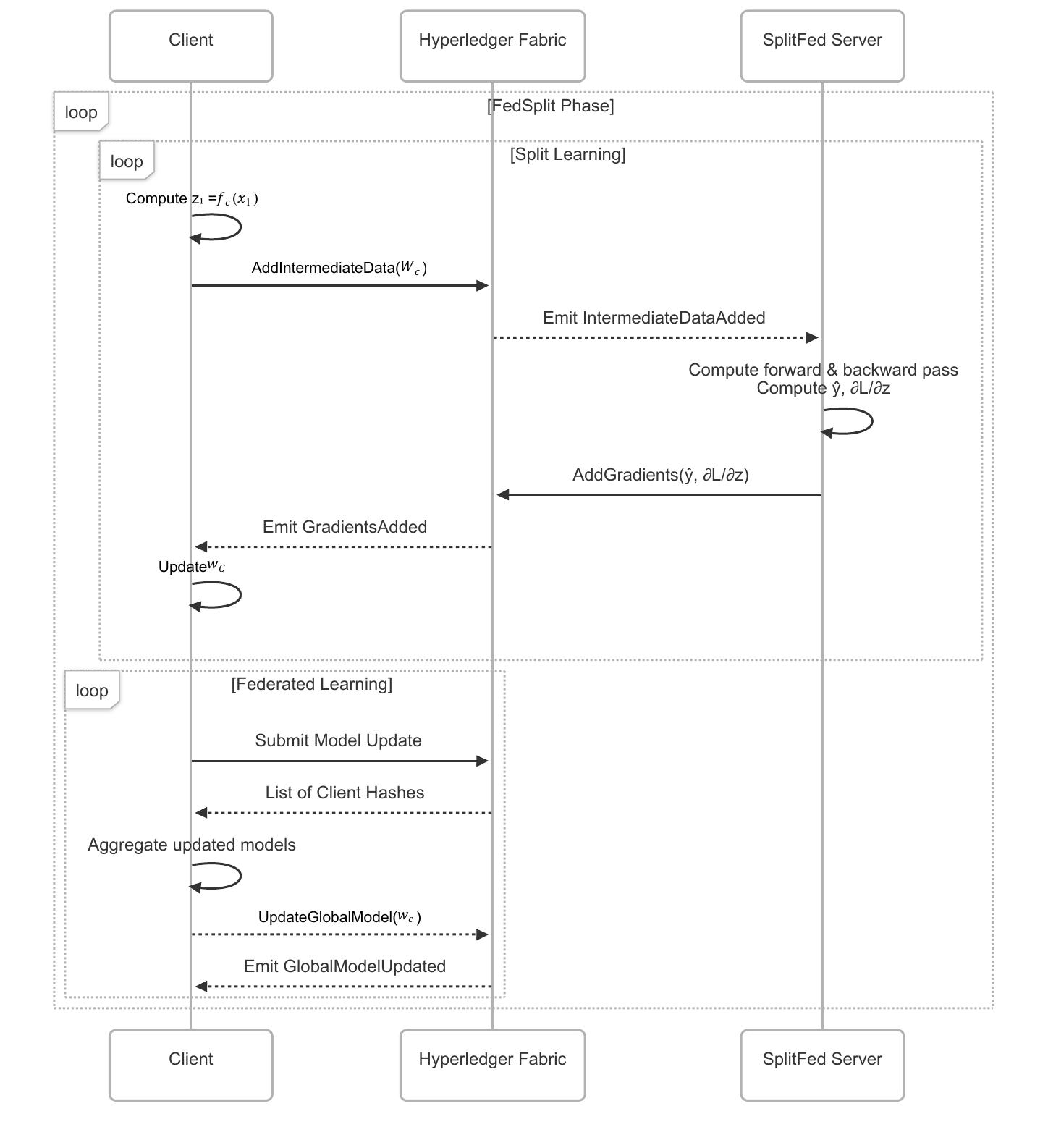}
  \caption{End-to-end training workflow showing iterative split learning phases (inner loop) and periodic federated aggregation (outer loop). Blue arrows represent blockchain transactions, gold arrows denote off-chain computations.}
  \label{fig:training-flow}
\end{figure*}

The collaborative training process (illustrated conceptually in Figs.~\ref{fig:model-lifecycle} and \ref{fig:training-flow}) operates in distinct phases, orchestrated via HLF transactions, chaincode logic managing PDC hash references, transient fields for data transfer, and events for coordination. The full end-to-end HLF-FSL training procedure is summarized in Algorithm~\ref{alg:workflow_hlf_fsl}.

\begin{algorithm}[tbp!]
\caption{HLF-FSL Training Workflow }
\label{alg:workflow_hlf_fsl} 
\begin{algorithmic}[1]
\State \textbf{Input:} Clients $\mathcal{C}$, Server Entity $\mathcal{S}$, Chaincode $\mathcal{CH}$, Off-chain Storage (OCS: IPFS/Client-Store)
\Procedure{Global Training Round}{$t$}
    \For{client $C_i \in \mathcal{C}$ \textbf{in parallel}}  \Comment{Split Learning Phase}
        \State $z_i \gets f_C(x_i; \mathbf{w}_{C}^{(t)})$ \Comment{Off-chain}
        \State $data\_ref\_z_i \gets C_i$ prepares $z_i$ for HLF (via OCS, gets hash/CID)
        \State $\mathcal{CH}.\texttt{addIntermediateData}(data\_ref\_z_i,\text{round}\ t, \text{metadata})$
    \EndFor \\

    \State $\mathcal{S}$ receives events, retrieves $\{data\_ref\_z_i\}$ from PDC A via $\mathcal{CH}$
    \State $\mathcal{S}$ fetches actual $\{z_i\}$ from Transient map (using TxID from event) or OCS (using hash)
    \State $\mathcal{S}$ computes $\{\hat{y}_i, \nabla_{\mathbf{z}_i} L_i\}$ and $\nabla_{\mathbf{w}_s}L_i$ off-chain
    \State $\mathcal{S}$ updates $\mathbf{w}_S^{(t+1)}$ off-chain
    \For{client $C_i \in \mathcal{C}$}
         \State $data\_ref\_\nabla z_i \gets \mathcal{S}$ prepares $\nabla_{\mathbf{z}_i} L_i$ for HLF (Transient/OCS, gets hash/CID)
        \State $\mathcal{S}$ submits $(C_i, data\_ref\_\nabla z_i)$ to $\mathcal{CH}.\texttt{addGradients}$ 
    \EndFor \\

    \For{client $C_i \in \mathcal{C}$ \textbf{in parallel}} \Comment{Local Update \& Parameter Hash Submission}
        \State $C_i$ receives event, retrieves $data\_ref\_\nabla z_i$ from PDC A via $\mathcal{CH}$
        \State $C_i$ fetches actual $\nabla_{\mathbf{z}_i} L_i$ from Transient map (via TxID) or OCS
        \State $\tilde{\mathbf{w}}_{C,i}^{(t+1)} \gets \texttt{SGD}(\mathbf{w}_{C}^{(t)}, \nabla_{\mathbf{z}_i} L_i)$ \Comment{Off-chain}
    \EndFor \\

    \If{aggregation triggered for round $t$ by $\mathcal{CH}$} \Comment{Aggregation Phase}
        \State $\mathcal{CH}.\texttt{triggerClientAggregation}(\text{aggregationID for round } t)$ \Comment{Emits event with list of $\{hash\_\tilde{w}_{C,j}\}$ from PDC B}
        \For{client $C_k \in \mathcal{C}$ \textbf{in parallel}} \Comment{Each participating client computes global model}
            \State $C_k$ receives event with list of $\{hash\_\tilde{w}_{C,j}\}$ for aggregation
            \State $C_k$ retrieves actual parameters $\{\tilde{\mathbf{w}}_{C,j}^{(t+1)}\}$ 
            \State $\bar{\mathbf{w}}_{C,k}^{(t+1)} \gets C_k$ computes its version of FedAvg (Eq. \ref{eq:fsl_client_agg}) off-chain
            \State $hash\_\bar{\mathbf{w}}_{C,k} \gets \text{Hash}(\bar{\mathbf{w}}_{C,k}^{(t+1)})$
            \State $\mathcal{CH}.\texttt{commitGlobalModelHash}(\text{aggregationID for round } t, hash\_\bar{\mathbf{w}}_{C,k})$ 
        \EndFor         
        \State Clients collaboratively agree on $hash\_\bar{\mathbf{w}}_C \gets \text{Hash}(\bar{\mathbf{w}}_C^{(t+1)})$
        \State $\mathbf{w}_C^{(t+1)} \gets \text{Reference to PDC C hash on ledger}$
    \Else
        \State $\mathbf{w}_C^{(t+1)} \gets \mathbf{w}_C^{(t)}$ \
    \EndIf
\EndProcedure
\end{algorithmic}
\end{algorithm}

\paragraph{Model Registration \& Discovery} Server Entities register their $f_s$ hosting capabilities via chaincode. Clients then publish model architectures ($f_c, f_s$), storing detailed definitions off-chain (e.g., in PDCs with restricted access or a shared repository) and their corresponding hashes and metadata on the main ledger via chaincode. Clients discover models by querying this on-chain registry and retrieve necessary components after Fabric's MSP verifies their authorization.

\paragraph{Split Learning Training Iterations}
Clients compute intermediate activations $z_i$ (Alg.~\ref{alg:workflow_hlf_fsl}, line 4). A verifiable reference to $z_i$ is submitted via chaincode (Fig.~\ref{fig:architecture}, step 1), which records this reference in the (`intermediateDataHashCollection`, policy: Client+Server+Admin) and emits an event(Alg.~\ref{alg:workflow_hlf_fsl}, lines 5-6). The Server Entity, notified by this event, retrieves the reference (Fig.~\ref{fig:architecture}, step 2), then fetches the actual data(Alg.~\ref{alg:workflow_hlf_fsl}, lines 9-10). After off-chain computation (Alg.~\ref{alg:workflow_hlf_fsl}, line 11-12), the Server Entity generates gradients $\nabla_{\mathbf{z}_i} L_i$ and sends them back to the client, with its reference recorded by chaincode in the same `intermediateDataHashCollection` PDC (Fig.~\ref{fig:architecture}, step 3)(Alg.~\ref{alg:workflow_hlf_fsl}, line 14-15). Clients retrieve these gradients, perform local backpropagation, and compute their updated local parameters $\tilde{\mathbf{w}}_{c,i}^{(t+1)}$ (Fig.~\ref{fig:architecture}, step 4) (Alg.~\ref{alg:workflow_hlf_fsl}, lines 19-21).

\paragraph{Federated Aggregation}
At predefined intervals, clients store their locally updated parameters $\tilde{\mathbf{w}}_{c,i}^{(t+1)}$ securely off-chain  and submit the hash of these parameters to the chaincode via \texttt{submitClientModelHash} (Fig.~\ref{fig:architecture}, step 5). These hashes are stored in a PDC (`clientModelHashCollection`, policy: Client+Admin), with references on the main ledger.
The chaincode logic (e.g., \texttt{triggerClientAggregation}) then identifies the set of submitted model parameter hashes for the current aggregation cycle and makes this list available to all participating clients, typically via a Fabric event (Fig.~\ref{fig:architecture}, step 6) (Alg.~\ref{alg:workflow_hlf_fsl}, line 25).
Upon receiving this list, clients collaboratively perform the FedAvg computation off-chain. This involves:
\begin{enumerate}[label=(\alph*)]
    \item Each client fetching the list of required parameter hashes.(Alg.~\ref{alg:workflow_hlf_fsl}, line 27)
    \item Clients retrieving the actual model parameters $\tilde{\mathbf{w}}_{c,j}^{(t+1)}$ corresponding to these hashes from the designated location, such as a Private Data Collection or a distributed file system like IPFS. (Alg.~\ref{alg:workflow_hlf_fsl}, line 28)
    \item Each client computing the new global model $\bar{\mathbf{w}}_c^{(t+1)}$ using FedAvg. (Alg.~\ref{alg:workflow_hlf_fsl}, line 29)
    \item Clients submitting their global model to the chaincode function \texttt{commitGlobalModelHash} (Fig.~\ref{fig:architecture}, step 7) (Alg.~\ref{alg:workflow_hlf_fsl}, line 30-31). This transaction must satisfy an endorsement policy requiring approval from multiple client organizations, ensuring collective agreement on the new global model.
\end{enumerate}

The chaincode verifies the submission and records the global model hash in the `globalModelHashCollection` PDC (policy: All Clients+Admin for read; write restricted by endorsement policy) (Alg.~\ref{alg:workflow_hlf_fsl}, line 33-34). A `GlobalModelUpdated` event then signals all clients to fetch the new global model using this committed hash. 

This event-driven, chaincode-governed workflow significantly decentralizes the aggregation process. The on-chain logic initiates and validates the aggregation, while clients collaboratively perform the computation, a process verified through HLF's endorsement policies. This leverages HLF for coordination and verification while clients manage the computational load.

 \subsection{Chaincode Design for Orchestration}
\label{subsec:chaincode_design}

The chaincode acts as the decentralized orchestrator, managing workflow state transitions, data references (hashes stored in PDCs pointing to transient, IPFS, or client-managed off-chain data), participant identities, and coordinating the client-executed aggregation.

\paragraph{Participant and Model Management} Functions such as \texttt{registerServer}, \texttt{registerClient}, \texttt{publishModel}, \texttt{getAvailableModels}, and \texttt{getModel} handle the onboarding of server entities and clients, and the registration and discovery of FSL model architectures. These functions primarily record metadata and cryptographic hashes on the main ledger, with PDCs used to store these hashes under specific access control policies if further restriction beyond channel membership is needed for model definitions.

\paragraph{Orchestrating Split Learning Iterations}
The \texttt{addIntermediateData( roundID, dataHash, metadataJson) } chaincode function is invoked by clients. It expects the actual sensitive activation data $z_i$ to have been passed via a transient field. If $z_i$ was too large for the transient field and stored on IPFS, `dataHash` would be its IPFS CID, also passed within the transient map for verification. The chaincode verifies the client, calculates a hash of the received transient data (or verifies the provided IPFS CID), stores this definitive hash in the `intermediateDataHashCollection` PDC (Policy: Client Org + Server Org + Admin Org), records associated metadata on the ledger, and emits an `IntermediateDataAdded` event containing this hash and the original transaction ID for the Server Entity.
Similarly, the \texttt{addGradients(clientID, roundID, gradHash, metadataJson)} function, invoked by the Server Entity, expects gradient data $\nabla_{\mathbf{z}_i} L_i$. It stores the verified hash in the same `intermediateDataHashCollection` PDC and emits a `GradientAdded` event with this hash and TxID for the specific client.

\paragraph{Client Update Submission and Aggregation Coordination}
After local updates, clients invoke \texttt{submitClientModelHash (roundID, aggregationID, modelParamHash, metadataJson)}. This function receives the hash of the client's off-chain stored parameters $\tilde{\mathbf{w}}_{c,i}^{(t+1)}$ (where parameters themselves are on client-controlled storage or IPFS) and stores this `modelParamHash` along with metadata (e.g., dataset size) in the `clientModelHashCollection` PDC (Policy: Submitting Client Org + Admin Org). A reference is also logged on the main ledger state.
The client-led aggregation is coordinated by:
\begin{itemize}
    \item \texttt{triggerClientAggregation(aggregationID)}: Invoked periodically or by a designated coordinator role, this function queries the main ledger for all submitted `modelParamHash` references linked to the `aggregationID` (which point to hashes in PDC B). It then emits an `AggregationTaskStart` event containing this list of model parameter hashes.
    \item \texttt{queryClientModelHashesForAggregation(aggregationID)}: An optional query function allowing clients to pull the list of model parameter hashes needed for the current aggregation cycle, if they missed the event or join mid-cycle.
    \item \texttt{commitGlobalModelHash(aggregationID, roundID, aggregatedGlobalModel)}: Invoked by each client (or aggregation participant) after it has independently computed its version of the global model. This function records the `aggregatedGlobalModel` submitted by the client, associating it with the `aggregationID` and the client's identity.
    \item \texttt{endGlobalModel(aggregationID, roundID)}: This critical function is invoked (potentially by the submit of the last client or after a timeout) to conclude the aggregation cycle. It reads all `aggregatedGlobalModel` submissions for the given `aggregationID`. It then applies a consensus rule (e.g., checks if more than two-thirds of participants made an identical submission). If consensus is met, it selects the agreed-upon upload, stores this `globalModelHash` in the `globalModelHashCollection` PDC (Policy: All Client Orgs + Admin Org for read; write restricted by this function's endorsement policy, e.g., requiring M-of-N client Org signatures), and updates the main ledger to point to this new global model reference. A `GlobalModelUpdated` event, containing the final `globalModelHash`, is then emitted to all clients. If consensus is not met, the chaincode might log a failure or revert to the previous round's global model. The endorsement policy for this function is critical, as it ensures that the commitment of a new global model is a collective, verifiable decision, thereby replacing the trust placed in a central aggregator.
\end{itemize}
This chaincode design ensures that HLF orchestrates the FSL process by managing verifiable references and state transitions, while sensitive data exchange and heavy computations remain off-ledger.

\subsection{Privacy Mechanisms}

\paragraph{Transient Field} A basis of our privacy architecture is the strategic use of HLF's transient fields. Intermediate activations ($z_i$) and gradients ($\nabla_{\mathbf{z}_i} L_i$), while essential for training, are highly sensitive and have been shown to be vulnerable to information leakage and reconstruction attacks \cite{pasquini2022eluding, abuadbba2020can}. To mitigate them, these artifacts are passed within the transient field of a transaction proposal. This makes the data available to peers for chaincode execution and validation but excludes it from being written to the transaction on the ledger. By preventing the persistence of this data, we address a privacy vulnerability of the SL paradigm while still allowing for a verifiable, on-chain record of the interaction itself.

\paragraph{Controlled Access to Data Hashes via Private Data Collections (PDCs)}
Rather than storing any raw ML data or full model parameters on the main ledger, our architecture utilizes PDCs to store only cryptographic hashes that reference the actual off-chain data. Specific PDCs are defined with granular access policies enforced by Fabric's MSP, ensuring need-to-know access:
    \begin{itemize}
        \item \texttt{intermediateDataHashCollection}: Stores hashes of $z_i$ and $\nabla_{\mathbf{z}_i} L_i$. Its policy (`OR('ClientMSP.member', 'ServerMSP.member', 'AdminMSP.member')`) grants access only to the involved Client, the Server Entity processing its data, and an Admin Org for audit, preventing other clients or unauthorized parties from accessing even the references to this intermediate data exchange.
        \item \texttt{clientModelHashCollection}: Stores hashes of individual client-updated parameters $\tilde{\mathbf{w}}_{c,i}^{(t+1)}$. The policy (`OR('SubmittingClientMSP.member', 'AdminMSP.member')`) restricts access to only the client organization that owns the update and the Admin Org. This is crucial as it prevents the Server Entity and other clients from directly accessing hashes of individual (non-aggregated) client models.
        \item \texttt{globalModelHashCollection}: Stores the hash of the aggregated global client-side model $\bar{\mathbf{w}}_c^{(t+1)}$. Its policy (`OR('AllParticipatingClientMSPs.member', 'AdminMSP.member')` for read access) allows all participating clients and the Admin Org to retrieve the reference to the latest global model. Write access is strictly governed by the `commitGlobalModelHash` chaincode function's endorsement policy, which requires consensus among client organizations.
    \end{itemize}
This strategy of storing only hashes in PDCs, coupled with precisely defined access policies, provides robust data obfuscation and ensures that even references to sensitive data are not universally exposed.

\section{Experimental and Evaluation Scenario}
\label{sec:experimental_setup}

This section details the experimental environment, datasets and machine learning models employed, baseline frameworks used for comparison, specific metrics collected and the distinct experimental scenarios designed to assess performance and scalability.

Our evaluation and simulation\footnote{Code is available online: \\ \textsf{\url{https://gitlab.com/compromise3/hlf-fsl}}} of the HLF‐FSL platform was carried out on a uniform hardware and software stack to isolate the impact of the blockchain integration. All machine learning components were implemented in PyTorch (v3.9) and executed on a workstation equipped with an Intel Core i9-12900K CPU, 32 GB DDR5 RAM, an RTX 3090 Ti GPU, and a 1 TB NVMe SSD. The Hyperledger Fabric network (v2.5 LTS) was provisioned via Docker Compose, employing the Raft consensus protocol and Go (v1.23.9) chaincode. Clients, the server entity, and the aggregation process communicated with Fabric through Python SDK wrappers exposing REST and WebSocket interfaces. 

For the learning tasks, we selected CIFAR-10  as our primary image classification benchmark \cite{CIFAR10} and, for an abroad comparison, the MNIST digit dataset \cite{deng2012mnist}. On CIFAR-10, we used a ResNet-18 \cite{he2016deep} architecture initialized from scratch, splitting the model after the first residual block (PyTorch’s `layer1`) so client‐side subnetwork is formed by the initial convolution, normalization, activation, pooling, and `layer1` layers, while the server‐side has the remaining blocks, pooling, and classifier head. 

For MNIST we employed a standard five‐layer CNN (two conv–pool blocks followed by two fully connected layers and softmax), split at the first pooling layer to mirror the ResNet-18 division. In all experiments, training proceeded for 50 global epochs with dynamic and variable hyperparameters depending of the characteristics of the scenario.

We evaluated both learning quality and system behavior using a unified set of metrics.  Model performance was measured by tracking global test accuracy and average test loss after each aggregation round and the evolution of training loss on the server‐side submodels.  To capture convergence speed, we measured the wall‐clock time and number of global epochs required to reach predefined accuracy thresholds. 

Inside the system, we measured the total epoch duration —client $f_c$ fw/bw passes, submission and retrieval of data hashes via Fabric transactions, server $f_s$ processing, and FedAvg aggregation. This metrics provide latency breakdown separating client computation, server computation, aggregation, and SDK call delays. Throughput was expressed both as HLF transaction throughput (TPS) and as the effective rate of client activations processed per second.  

Communication costs were estimated by the serialized byte sizes of activations, gradients, and client/server model parameters exchanged. We approximated ledger growth by measuring the cumulative on-chain metadata and PDC reference sizes incurred for every training round. All timing measurements used wall-clock timers synchronized across components, with metrics captured automatically by our Python implementation.

To contextualize the performance of HLF‐FSL, we compared it against three baselines implemented on the same hardware. First, a standalone FSL Python orchestrator managing a central server and FedAvg aggregation. Second, a standalone SL. Third, our prior SL+Ethereum implementation \cite{beis24}, where SL ran over an Ethereum smart‐contract backbone. Each baseline adhered to identical model splits, datasets, and hyperparameters to ensure a fair comparison.

Our experimental suite comprised four scenarios: (1) functional validation under IID data on CIFAR-10 and MNIST with \(N=10\) clients to establish baseline accuracy and epoch time; (2) robustness to statistical heterogeneity using Dirichlet‐simulated non-IID partitions, (3) scalability tests varying the number of clients to assess parallelism gains and overheads analysis and (4) direct platform comparison between HLF‐FSL and other State-of-the-Art methods on identical tasks Experiments involving randomness (e.g., data splits, weight initializations) were repeated to report mean and standard deviation, ensuring statistical rigor.

\section{Results}
\label{sec:results}

\subsection{Functional Validation \& Baseline Performance }
\label{subsec:results_EXP1}

The first experiment aims to validate the correct behavior of the platform and evaluate its baseline machine learning results across our two benchmark datasets: CIFAR-10 and MNIST.

\subsubsection{CIFAR-10 Benchmark Results}

Table~\ref{tab:exp1_summary} shows the final test accuracy and average per-epoch training time. The models trained successfully converged, achieving a final mean test accuracy of 94.14\%. This result is comparable to the centralized FSL baseline, which reached 94.7\%, indicating that the cost of introducing HLF has minimal impact on final learning performance, as it should be. Compared to sequential SL approaches, it shows improved accuracy; the centralized SL baseline achieved 91.2\%, and our previous SL-Ethereum implementation \cite{beis24} averaged 90.25\% final accuracy across its clients. This suggests a learning benefit from FSL's parallel client processing and aggregation strategy over purely sequential SL.

\begin{table}[tbp!] 
  \centering
  \caption{EXP 1: Performance Summary (CIFAR-10, IID, $N=10$ Clients, 50 Epochs)}
  \label{tab:exp1_summary}
  \begin{tabular}{@{}lcc@{}}
    \toprule
    \textbf{Method} & \textbf{Final Test Acc. (\%)} & \textbf{Avg. Epoch Time} \\
    \midrule
    \textbf{HLF-FSL } & \textbf{94.14} & \textbf{30m 38s} \\ 
    SL-Ethereum (Avg. from \cite{beis24}) & 90.25 & 1h 25m \\ 
    Standalone FSL (In-Memory) & 94.7 & 5m 50s \\ 
    Standalone SL (In-Memory)  & 91.2 & 5m 10s \\ 
    \bottomrule
  \end{tabular}
\end{table}

In terms of execution time, our work recorded an average epoch time of approximately 30 minutes and 38 seconds under 10 clients and 64 images for training batch size. This represents an improvement compared to the Ethereum implementation, which averaged 1 hour and 25 minutes for every epoch in a similar machine learning task. The centralized FSL (5m 50s) and centralized SL (5m 10s) baselines, operating in-memory, have faster epoch times due to the absence of blockchain and communication overhead. The overhead introduced comes from SDK interactions and transaction processing (endorsement, ordering, validation).

\begin{figure*}[tbp!]
    \centering
    \begin{subfigure}[b]{0.48\textwidth}
        \centering
        \includegraphics[width=\linewidth]{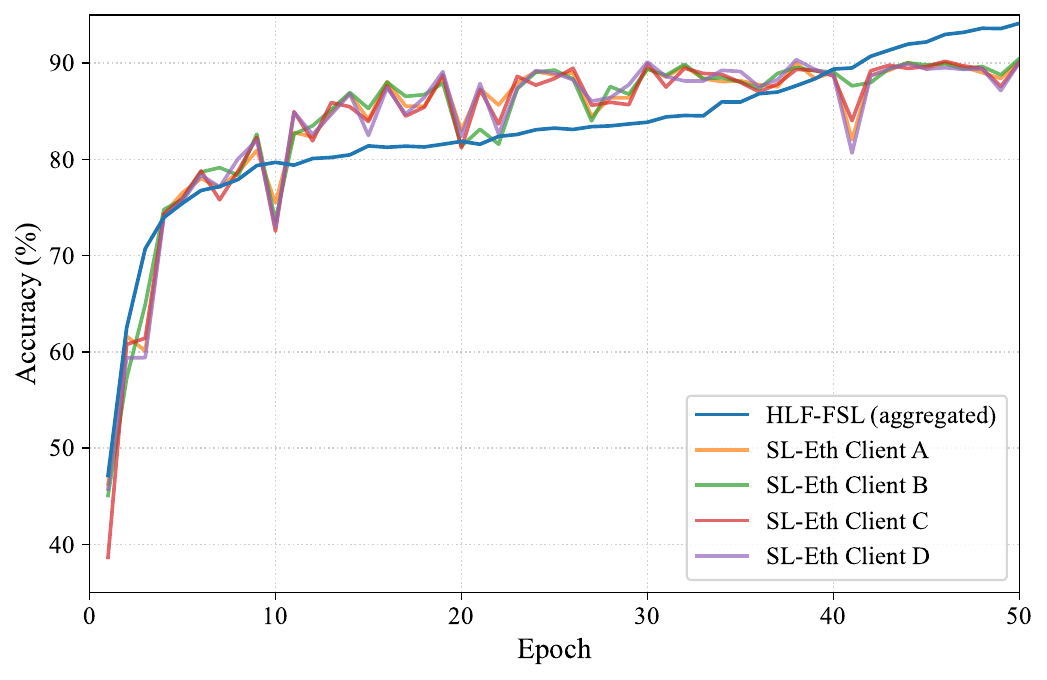}
        \caption{}
        \label{fig:exp1_acc_methods}
    \end{subfigure}
    \hfill
    \begin{subfigure}[b]{0.48\textwidth}
        \centering
        \includegraphics[width=\linewidth]{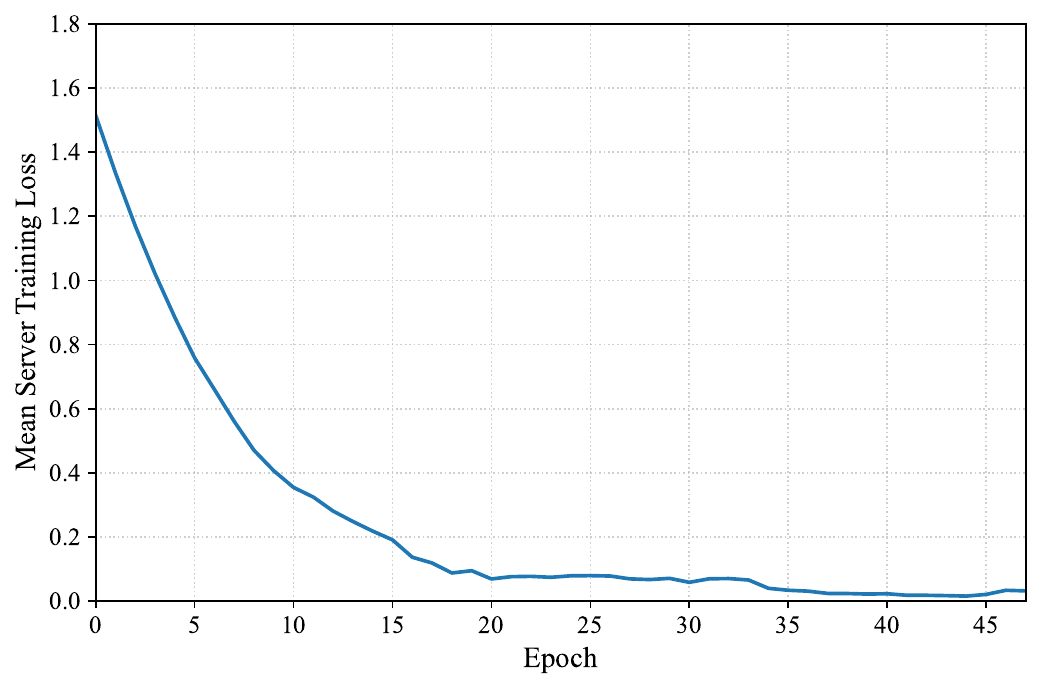}
        \caption{}
        \label{fig:exp1_loss_methods}
    \end{subfigure}
    \caption{a) Test accuracy vs.\ global epoch for HLF-FSL (aggregated global model, $N=10$ clients) compared to individual client training accuracies from the SL-Ethereum system. b) Test Loss vs. Epoch }
    \label{fig:exp1_convergence_methods}
\end{figure*}

Figure~\ref{fig:exp1_acc_methods} showcases the convergence of the global model, contrasting it with the individual client learning curves from the SL-Ethereum system. Figure~\ref{fig:exp1_loss_methods} shows the corresponding test loss versus epoch, highlighting constant decline in loss until convergence by 40-50 epoch.

\subsubsection{MINST Benchmark Results}

To facilitate comparison with a broader range of works, including scenarios where it is common work with simpler models, we evaluate the platform also on the MNIST dataset. As detailed in Table~\ref{tab:exp1_mnist_summary}, the architecture shows exceptional performance on MNIST, achieving an accuracy of 99.43\%, being the average training time for epoch 7 minutes and 20 seconds. The global model reached 98\% accuracy in the first 5 epochs and stabilized near its peak performance. 

This results underscore the practical viability and capability of the proposal to host machine learning tasks while operating within a decentralized and verifiable blockchain framework.

\begin{table}[tbp!] 
  \centering
  \caption{MNIST Performance Summary}
  \label{tab:exp1_mnist_summary}
  \begin{tabular}{@{}lc@{}}
    \toprule
    \textbf{Metric} & \textbf{Value} \\
    \midrule
    Final Test Accuracy (\text{Epoch 50}) & 99.43\% \\
    Average Epoch Time & 7m 20s \\
    Peak Accuracy (Epoch 38) & 99.48\% \\
    Epochs to Reach 98\% Accuracy & 5 \\
    \bottomrule
  \end{tabular}
\end{table}

\begin{figure}[tbp]
\centering
\includegraphics[width=0.45\linewidth]{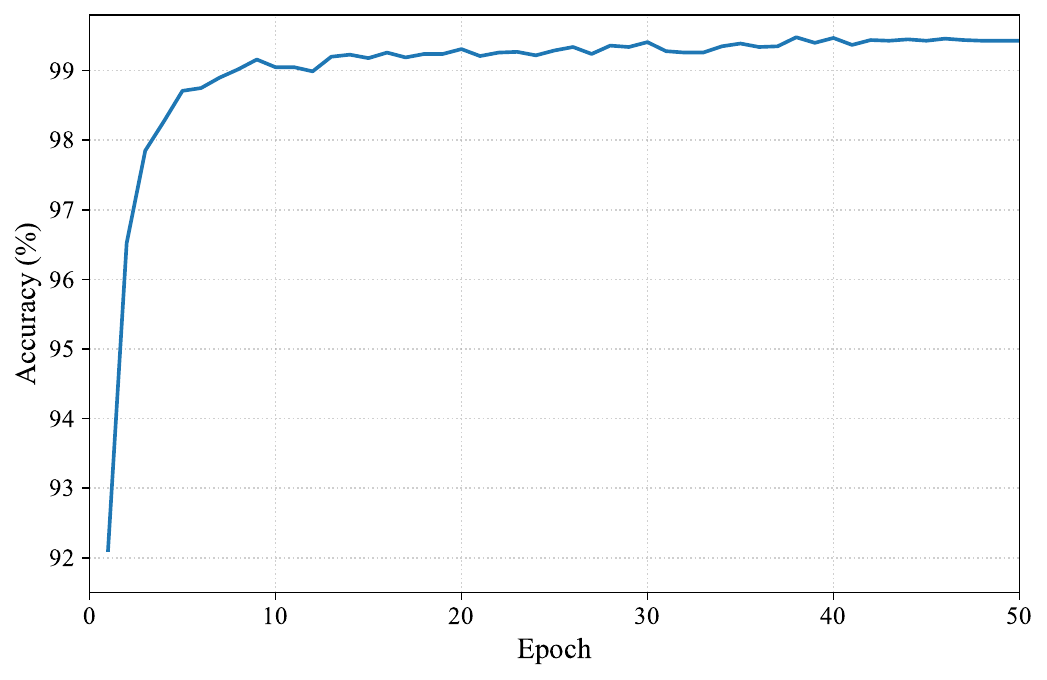}
\caption{MNIST: HLF-FSL global model test accuracy convergence over epochs.}
\label{fig:exp1_mnist}
\end{figure}

\subsection{Performance under Non-IID Conditions}
\label{subsec:results_non-iid}

A critical challenge in real-world collaborative learning is statistical heterogeneity, where client data is not Independent and Identically Distributed (Non-IID). We assessed the robustness of our HLF-FSL platform to such conditions using the CIFAR-10 dataset, partitioned among $N=10$ clients via a Dirichlet distribution with varying concentration parameters ($\alpha \in \{0.1, 0.5, 0.9\}$). A smaller $\alpha$ indicates a higher degree of data skew. The ResNet-18 model, split point, and training hyperparameters were consistent with the IID experiments.

\begin{figure*}[tbp!]
    \centering
    \begin{subfigure}[b]{0.48\textwidth}
  \centering
  \includegraphics[width=\linewidth]{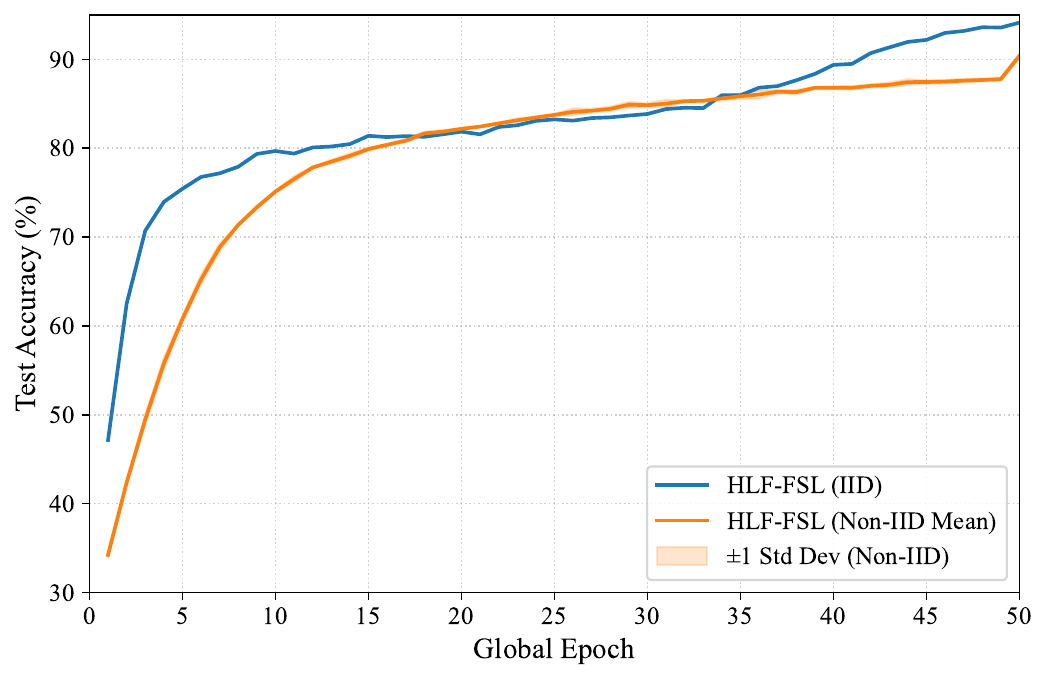} 
  \caption{}
  \label{fig:exp3_accuracy}
    \end{subfigure}
    \hfill
    \begin{subfigure}[b]{0.48\textwidth}
   \centering
   \includegraphics[width=\linewidth]{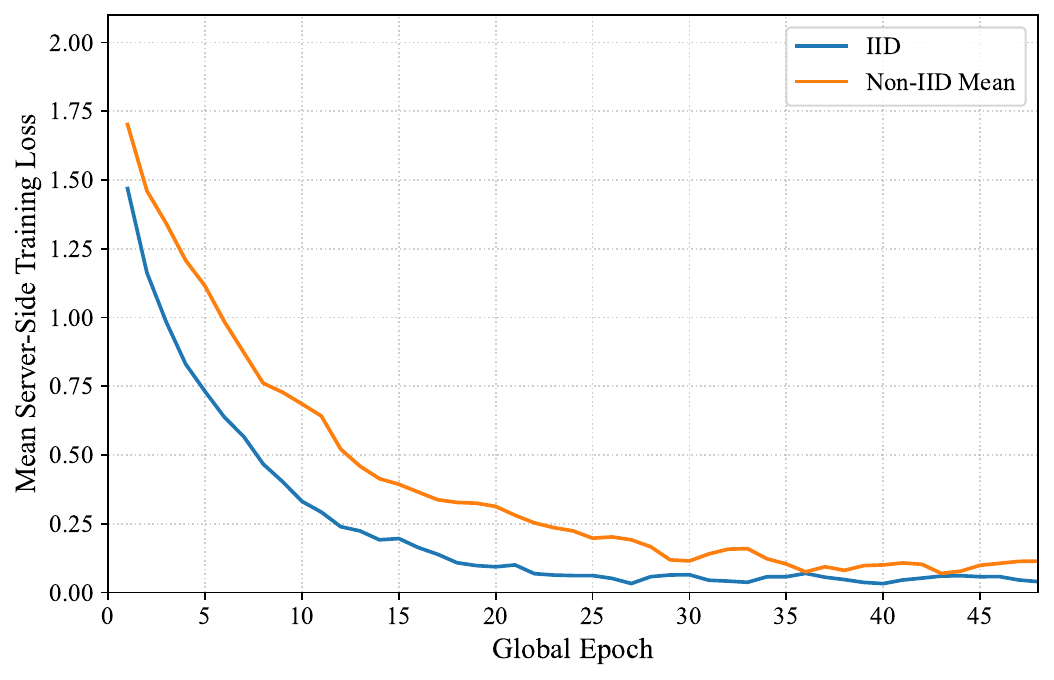}
   \caption{}
   \label{fig:exp3_loss}
    \end{subfigure}
    \caption{ Non-IID Robustness: Accuracy and Loss convergence of the HLF-FSL global model under IID conditions and Non-IID conditions simulated with Dirichlet distribution parameters $\alpha \in \{0.1, 0.3, 0.5, 0.9\}$.}
    \label{fig:exp1_iid_vs_niid}
\end{figure*}

Figures~\ref{fig:exp3_accuracy} and \ref{fig:exp3_loss} presents the accuracy and loss convergence of the HLF-FSL global model for both the IID scenario (replicated from EXP 1 for clarity) and the Non-IID (the represented values are a mean of the values from $\alpha$ 0.1, 0.3, 0.5 and 0.9) scenario. The platform demonstrates strong learning performance even under significant data heterogeneity. While, as expected, the final accuracy under Non-IID conditions is lower than in the IID case, the degradation is relatively graceful. These results are notably resilient, especially considering the challenging nature of CIFAR-10 under such distributions. The convergence curves for Non-IID setting closely track each other and the IID case for the middle epochs, diverging slightly at the initial and end of the training but still achieving high accuracies. This suggests that the FSL paradigm, with its client-side feature extraction ($f_c$) and subsequent federated averaging of these representations, coupled with the stable learning of the shared server-side model ($f_s$), provides a commendable degree of intrinsic robustness to the class imbalances introduced by Dirichlet partitioning. The decentralized aggregation mechanism coordinated by HLF appears effective in mitigating the negative impacts of individual client data skew on the global model's generalization capability.

\subsection{Scalability Analysis }
\label{subsec:results_scalability}

Understanding the computational and communication overhead introduced by the architecture is crucial to validate its practical viability. This section dissects these overheads by analyzing detailed timing and data volume metrics collected during every epoch of training with different numbers of clients. 

\subsubsection{Off-Chain Computational Overhead}
The primary off-chain computations involve the client-side forward ($f_c$) and backward passes, the server entity's batched processing of activations through $f_s$, and the aggregator's federated averaging.

Table~\ref{tab:exp2_compute_overhead} presents the mean duration for these components, averaged across clients and epochs for different configurations.

\begin{table}[tbp!]
\centering
\caption{Mean Off-Chain Computational Times (seconds) per Component}
\label{tab:exp2_compute_overhead}
\sisetup{table-format=2.4}
\begin{tabular}{@{}S[table-format=2.0] SSSS@{}} 
\toprule
{\textbf{($N$ Clients)}} & {\textbf{Client $f_c$ Fwd}} & {\textbf{Client $f_c$ Bwd}} & {\textbf{Server $f_s$ Batch}} & {\textbf{Aggregator Compute}} \\
\midrule
1    & 0.0038 & 0.0046 &  0.6040 & 0.1456 \\
2    & 0.0252 & 0.0612 &  0.9071 & 0.1306 \\
10   & 0.0843 & 0.0878 &  3.2471 & 0.1593 \\
15   & 0.1011 & 0.0803 &  1.8504 & 0.2464 \\
20   & 0.0812 & 0.0851 & 13.4787 & 0.1691 \\ 
25   & 0.1380 & 0.1179 & 26.3214 & 0.2997 \\ 
\bottomrule
\end{tabular}
\end{table}

As observed in Table~\ref{tab:exp2_compute_overhead}, the client-side forward and backward pass times per batch remain relatively low and stable, reflecting the efficiency of the $f_c$ portion of the ResNet-18 model. The `Server $f_s$ Batch` processing time shows more variability and increases with the number of clients, as the server entity processes activations from more clients, even with batching. For $N=20, B=64$, the server batch processing averaged $\approx$13.48s, and for $N=25, B=64$, it increased to $\approx 26.32$ s. This indicates that the server entity's capacity to concurrently handle and batch-process incoming activations is a key factor in overall system throughput, especially as client numbers scale. The `AggregatorCompute` time, representing the off-chain FedAvg computation, remains consistently low, as it primarily involves averaging model parameters whose structure is fixed.

\subsubsection{Communication and SDK Interaction Overhead}
Communication in HLF-FSL involves entities interacting with the Fabric network via SDK to invoke chaincode functions. Primary data (activations, gradients, model parameters) is handled off-chain, with only hashes/references and metadata stored on-chain.
Table~\ref{tab:exp2_sdk_latency} summarizes the mean latency for key SDK calls and communication volumes per transaction.

\begin{table*}[tbp!]
\centering
\caption{ Mean SDK Call Latency (seconds) and Communication Volume per Transaction.}
\label{tab:exp2_sdk_latency}
\resizebox{\textwidth}{!}{
\begin{tabular}{@{}lccccc@{}}
\toprule
\textbf{N Clients} & \textbf{`/addIntermediateData` (s)} & \textbf{`/addGradients`(s)} & \textbf{Avg. $z_i$ Size (MB)} & \textbf{ Client Model Upload Size (KB) } \\
\midrule
1   & 4.205 & 4.207 & $\approx$48.96 & $\approx$297 \\ 
2    & 4.180 & 3.974 & $\approx$48.95 & $\approx$297 \\ 
10   & 3.418 & 3.487 & $\approx$48.46 & $\approx$297 \\ 
15   & 3.601 & 3.495 & $\approx$48.15 & $\approx$297 \\ 
20  & 3.292 & 3.438 & $\approx$47.85 & $\approx$297 \\ 
25   & 3.070 & 3.490 & $\approx$48.99 & $\approx$297 \\ 
\bottomrule
\end{tabular}
}
\end{table*}

The SDK call latencies for `/addIntermediateData` (client submitting activation hash) and `/addGradients` (server submitting gradient hash) consistently average between 3-4.5 seconds across different client numbers. This duration includes network latency to the SDK wrapper, wrapper processing, Fabric SDK interaction with the peer, chaincode execution, endorsement, and submission to the ordering service. While not purely HLF transaction time, it represents the effective per-interaction overhead experienced by the off-chain applications. For $N=25$, latencies remain stable, indicating robust Fabric network performance despite the use of two terminals.

Activation volumes ($z_i$) are substantial (around 48–49 MB per transaction), underscoring the need for off-chain storage to avoid ledger bloat. Client model parameter updates ($\tilde{w}_{c,i}$), representing the first five layers of ResNet-18, are ~297 KB, making their hashes manageable on-chain and parameters suitable for PDC or off-chain transfer.

\subsubsection{Blockchain Ledger Overhead}
In our HLF-FSL design, the main channel ledger primarily stores metadata, event payloads, and cryptographic hashes or references (e.g., IPFS CIDs, PDC keys) to the actual large data objects (activations, gradients, model parameters) which reside off-chain or in transient memory/PDCs.
For instance, the analysis of the N=10 run indicated that the data passed to chaincode events (which gets recorded on the ledger) involved storing 46-byte CID strings for each intermediate data and gradient event. For 1580 such events of each type in every epoch 
\begin{equation*}
\text{Number of events} = \frac{50000 \times 2}{\text{batch\_size}} = \frac{50000 \times 2}{64} = 1580,
\end{equation*}
the total on-chain data specifically for these references was approximately 145 KB for one epoch. This strategy of keeping large data off-ledger and only managing small, verifiable references on-chain is key to maintaining a lean and efficient blockchain, minimizing storage growth and replication costs for participating peers.

This overhead analysis indicates that while the HLF integration introduces measurable latencies for on-chain interactions, the primary computational load remains off-chain. The strategy of using PDC-managed hashes effectively minimizes the on-chain storage footprint. The server entity's capacity for parallel batch processing and the efficiency of the underlying Fabric network's transaction handling are key determinants of overall platform scalability.

\subsection{Performace Comparison with State-of-the-Art Methods.}
\label{subsec:exp4_results}

This experiment compares the performance and characteristics of HLF-FSL against Ethereum-based distributed learning systems, specifically our previous work Split Learning + Ethereum~\cite{beis24} and other proposal in the state-of-the-art: the FSL-like Block-FeST framework~\cite{Batool2024}. The goal is size the advantage of implement a permissioned, enterprise-grade blockchain for FSL.

\subsubsection{HLF-FSL vs. SL+Ethereum}

Both systems were evaluated on CIFAR-10 using the Resnet-18 model, split after the residual block, with $N=10$ clients and training for 50 global epochs with a batch size of 64.

As we already see in Table\ref{tab:exp1_summary}, we observe substantial improvements in both learning and operational metrics. HLF-FSL achieved exceeding metrics both for accuracy, 94.14 vs 90.25, and training time, 30m 38s vs 1h 25m.

The accuracy gains are attributable to the enhanced metrics of FSL over SL. On the other hand, the operational gains are attributable to HLFc's architectural advantages over Ehtereum for iterative ML tasks. Raft consensus mechanism and faster transaction finality reduce the coordination overhead per round vs Ehtereum's Proof-of-Work, which incurs in high confirmation latencies and computational costs. Also, HLF built-in features like MSP or PDC provide a more robust and governable environment for collaboration, benefits not available in public Ethereum-based implementations.

\subsubsection{HLF-FSL vs. Block-FeST }

Block-FeST is one of the few existing frameworks that explicitly combines FSL with a blockchain backend (Ethereum in this case). While the application domain and model architecture differ from our image classification task with ResNet-18, a comparison of platform-level characteristics and design philosophies is insightful.

HLF-FSL establishes a fully peer trust model. All organizations (Clients, Server Entity, Admin) are authenticated via MSP. Chaincode, governed by endorsement policies, orchestrates the workflow, including triggering the client-executed aggregation. This contrasts with Block-FeST's description, which, while using Ethereum for an audit trail and smart contracts for orchestration, still implies distinct roles for a "Federated Learning server" and a "Split Learning server" that appear to manage core aspects of model processing and potential aggregation centrally before interacting with the blockchain. Our client aggregation decentralizes control compared to architectures relying on a server for the aggregation, even if that server logs results on a blockchain. Fabric's channels and PDCs offer data isolation and privacy controls for shared data, which are more granular than what is typically available on a public Ethereum ledger where data, once on-chain (even if hashed), is universally visible. 

While Block-FeST aims to offload computation from clients to an SL server, our HLF-FSL also achieves this client-side computational relief via the $f_c/f_s$ split. The key difference is that our HLF backbone for coordinating these interactions and the subsequent aggregation is inherently more performant and cost-effective than a public PoW blockchain. The communication overhead in HLF-FSL is managed by storing only compact hashes/references in PDCs and minimizing on-chain data footprint on the main ledger. \\

Unlike prior systems such as Block-FeST or our earlier SL+Ethereum, which primarily utilize public blockchains and often retain centralized elements for core FSL operations (like server-side model hosting or aggregation), our HLF-FSL platform achieves a higher degree of operational decentralization with significantly reduced blockchain-induced overhead. By strategically employing HLF's specific features—efficient consensus, PDCs for controlled hash storage and chaincode for verifiable orchestration of client-led aggregation—our prototype delivers faster training epochs, avoids direct transaction fees, and natively supports robust enterprise governance policies. These combined strengths make the HLF-FSL platform uniquely positioned for secure, efficient, and compliant cross-organizational collaborative learning in regulated settings, advancing beyond the limitations of previous FSL and blockchain-ML integrations.

\section{Conclusion and Future Work}
\label{sec:conclusion}

When we started this work, we did it with the aim of providing scalable, decentralized, and privacy-preserving collaborative machine learning suitable for sensitive environments. Our experimental evaluations demonstrated that HLF-FSL achieves robust performance, comparable accuracy to centralized FSL, and significant reductions in per-epoch training time compared to prior Ethereum-based implementations. By applying HLF's features we ensured secure and private handling of intermediate computations and model updates without sacrificing scalability or learning performance.

On CIFAR-10, we achieved 94.14\% accuracy, comparable to centralized FSL baselines (94.7\%), showing minimal performance impact from blockchain integration. It significantly reduced epoch training time to approximately 30 minutes and 38 seconds, a substantial improvement over the 1 hour and 25 minutes observed with prior Ethereum implementations and it also demonstrated robustness under non-IID data conditions, showing slightly accuracy degradation. Scalability analysis revealed minimal blockchain overhead, with most computational load remaining efficiently off-chain and the use of Private Data Collections (PDCs) effectively minimized on-chain storage overhead and allow to keep the data privacy while maintaining and immutable audit-trail. Compared to other state-of-the-art methods, HLF-FSL establishes a fully peer-trust model with chaincode orchestrating client-executed aggregation, offering more decentralized control and granular data isolation through Fabric's channels and PDCs than public Ethereum-based solutions. These combined strengths, including avoiding direct transaction fees, position it as a secure, efficient, and compliant solution tailored for cross-organizational collaborative learning in regulated enterprise settings.

The integration presented here not only demonstrates the feasibility of enterprise-grade decentralized federated split learning but also personally validates our initial hypothesis: blockchain technology, when properly tailored, can effectively decentralize trust and coordination, significantly mitigating privacy and scalability challenges inherent in traditional collaborative learning frameworks.

Future research directions for the proposed HLF-FSL framework include exploring the integration of robust decentralized aggregation methods, leveraging HLF’s endorsement policies for stronger computation correctness guarantees and fault tolerance. Incorporating advanced privacy mechanisms such as Differential Privacy (DP) and Homomorphic Encryption (HE) within the existing blockchain workflow could provide enhanced data confidentiality or embedded anomaly detection systems within Fabric's chaincode could proactively identify malicious updates, increasing overall system security. 

Additionally, extending evaluations to include detailed profiling of resource consumption (CPU, memory, bandwidth) under realistic enterprise workloads could optimize platform efficiency. Finally, validating the platform’s applicability in broader machine learning scenarios—such as natural language processing or object detection—and deploying it across geographically dispersed infrastructures would solidify the framework’s readiness for large-scale real-world adoption.

\appendices

\section*{Acknowledgment}
This work was supported by the grant PID2020-113795RB-C33 funded by MICIU/AEI/10.13039/501100011033 
COMPROMISE project), the grant PID2023-148716OB-C31 funded by MCIU/AEI/10.13039/501100011033 (DISCOVERY project); and “TRUFFLES: TRUsted Framework for Federated LEarning Systems, within the strategic cybersecurity projects (INCIBE, Spain), funded by the Recovery, Transformation and Resilience Plan (European Union, Next Generation)”. Additionally, it also has been funded by the Galician Regional Government under project ED431B
2024/41 (GPC).

\ifCLASSOPTIONcaptionsoff
  \newpage
\fi

\bibliographystyle{IEEEtran}

\end{document}